\documentclass[sn-basic]{sn-jnl}

\usepackage{graphicx}%
\usepackage{multirow}%
\usepackage{amsmath,amssymb,amsfonts}%
\usepackage{amsthm}%
\usepackage{mathrsfs}%
\usepackage[title]{appendix}%
\usepackage{xcolor}%
\usepackage{textcomp}%
\usepackage{manyfoot}%
\usepackage{booktabs}%
\usepackage{algorithm}%
\usepackage{algorithmicx}%
\usepackage{algpseudocode}%
\usepackage{listings}%

\theoremstyle{thmstyleone}%

\theoremstyle{thmstyletwo}%

\theoremstyle{thmstylethree}%

\begin{document}

\title{Adaptive Rigor in AI System Evaluation using Temperature-Controlled Verdict Aggregation via Generalized Power Mean}

\author[1]{\fnm{Aleksandr} \sur{Meshkov}}\email{alekslynx90@gmail.com}

\affil[1]{\orgname{Independent Researcher}, \orgaddress{\city{Budva}, \country{Montenegro}}}

\abstract{Existing evaluation methods for LLM-based AI systems, such as LLM-as-a-Judge, verdict systems, and NLI, do not always align well with human assessment because they cannot adapt their strictness to the application domain. This paper presents Temperature-Controlled Verdict Aggregation (TCVA), a method that combines a five-level verdict-scoring system with generalized power-mean aggregation and an intuitive temperature parameter $T \in [0.1, 1.0]$ to control evaluation rigor. Low temperatures yield pessimistic scores suited for safety-critical domains; high temperatures produce lenient scores appropriate for conversational AI. Experimental evaluation on three benchmark datasets with human Likert-scale annotations (SummEval and USR) shows that TCVA achieves correlation with human judgments comparable to RAGAS on faithfulness  (Spearman's $\rho = 0.667$ vs.\ $0.676$) while consistently outperforming DeepEval. The method requires no additional LLM calls when adjusting the temperature parameter.}


\maketitle

\section{Introduction}

In recent years, systems based on generative artificial intelligence (hereinafter referred to as AI) have received significant development and application in various fields of activity. Nowadays, in almost every field, medicine, finance, education, internal corporate processes, systems based on generative artificial intelligence are being implemented using technologies such as RAG and AI agents. Taking into account the availability of these technologies, the speed of development of new approaches to implementation, the emergence of systems that allow you to create your own AI agent with minimal coding skills, an important question arises: how to correctly evaluate the quality and reliability of such AI systems?

On the one hand, a fairly large number of benchmarks have recently appeared that allow to evaluate one or another aspect of an AI-based solution \cite{White2024,Ivanov2024benchmarks,Wang2024usercentric}, but if we are talking about an adaptive AI system that was created within a company to work with certain, non-public data, or an AI system that specializes in providing certain narrowly focused services for clients, then the use of benchmarks becomes irrelevant, since test datasets are more focused on testing the general capabilities of LLM models.

Unlike machine learning and deep learning neural networks, where to evaluate accuracy you can use part of the original dataset, which contains certain expected results for comparison, generative models do not provide a clear, unambiguous answer, and the same query can provide several possible answers, each of which may be correct and suitable from the point of view of human assessment. This leads to difficulties in evaluating AI systems based on generative models, but with the development of generative AI, solutions have appeared that allow such evaluation, for example, the DeepEval or RAGAS \cite{Es2023} libraries, which contain algorithms for evaluating metrics such as response relevance or identification of hallucinations and others.

However, during the practical use of these frameworks on various projects, it was revealed that the results of evaluating metrics do not always correlate well enough with human assessment. Often, such frameworks can underestimate the final score of a metric, despite the fact that a human assessment considers this answer to be quite relevant and correct.

This problem is the key topic to which this work is devoted, and is due to the fact that existing algorithms for evaluating AI systems, which will be further discussed in this work, are not able to adapt to the field of application of the AI system, which is why there is a discrepancy between the evaluation obtained from the framework and the human assessment.

Let's look at the following options for using AI systems. As a first example, it is worth considering the use of AI in medicine. To do this, let's imagine an AI system that can analyze a patient's symptoms and suggest possible diagnoses for treatment. When evaluating this AI system built on the basis of LLM models, it is important to understand that even partially incorrect information provided to the client can have critical consequences. For example, an incorrect diagnosis or treatment methods can threaten the life and health of the patient, therefore the evaluation of such AI systems should be as strict as possible and even one incorrect phrase or sentence in the response of the AI system, which may be caused by a hallucination or incorrect context, should immediately significantly reduce the final score of the metric being measured.

On the other hand, if we consider a regular AI chatbot designed to maintain a conversation with the user, then the appearance of a small hallucination in the answer, which does not contradict the facts, can be considered a correct answer. Users in this case may, on the contrary, value the chatbot's ability to improvise or maintain a lively dialogue, because the task of an AI chatbot in the current context is not absolute accuracy, but the overall satisfaction of users, therefore, the evaluation of such AI chatbots should be less strict.

In connection with the above, the question arose whether it is possible to adapt the evaluation metrics for LLM-based systems depending on the type of project and the requirements of human evaluation. As a solution to this issue, this work proposes a new algorithm for evaluating metrics for systems using generative AI, based on an extended version of the LLM as a Judge approach through a verdict system, but with the ability to control the severity of the evaluation through an additional temperature parameter.

The proposed algorithm introduces three key changes to the standard verdict-based evaluation pipeline:
\begin{enumerate}
   \item A five-level verdict system (instead of binary or ternary) that captures finer distinctions in the degree of compliance of AI system responses with evaluation criteria.
   \item Aggregation of verdict weights via the generalized power mean \cite{Hwang1981}, whose exponent parameter $p$ controls how strongly low verdicts affect the final score.
   \item An intuitive temperature parameter $T \in [0.1, 1.0]$ mapped to $p$, where low temperature produces strict evaluation and high temperature produces lenient evaluation, making the method accessible to practitioners without requiring knowledge of the underlying mathematics.
\end{enumerate}

\section{Review of existing evaluation approaches}

Before considering the proposed method, it is necessary to analyze existing approaches, determine their advantages and disadvantages in order to understand how the proposed solution reuses these approaches to evaluate AI systems based on LLM and what is the exclusive feature of the new proposed approach. To do this, the following approaches will be analyzed: LLM as a Judge with a direct request, LLM as a Judge with a verdict system, and approaches based on NLI.

\subsection{LLM as a Judge with a simple prompt}

The LLM as a Judge approach \cite{Zheng2023} is based on the idea that if LLM generation algorithms are able to generate answers that are close to human judgments, then it is possible to use these models to evaluate other models. This approach has become quite widespread in terms of the use of a large number of modern frameworks for evaluating AI-based systems.

A simple prompt means that the specialist performing the evaluation creates a prompt in which they indicate the data for evaluation, as well as the evaluation criteria. As a result, the LLM model evaluates the data based on the specified evaluation criteria in the provided prompt and assigns a final score.

An evaluation prompt might look like this: 

\begin{quote}
\textit{Rate the quality of the following answer to the question.} 

\textit{Question: What is the capital of France?} 

\textit{Answer: Paris is the capital of France, known for its art and culture.} 

\textit{Rate the relevance of the answer on a scale of 0 to 10, where:} 
\begin{itemize} 
\item \textit{0 means completely incorrect or irrelevant} 
\item \textit{10 means perfect, comprehensive answer} 
\end{itemize} 

\textit{Provide a single number rating.} 

\textit{Rating: \_\_} 
\end{quote}

As a result, LLM generates a score, for example 8, which shows how well the specified data meets the evaluation criteria.

Unfortunately, this approach has a significant drawback, namely, in most cases, a utility bias, since the model tries to be useful to the user, which leads to some overestimation, so even if the answer is not completely relevant or contains unnecessary information, the model will still give a high score to the metric.

To optimize this approach, a number of methods have been developed that improve the quality of evaluation through the use of large language models: 
\begin{enumerate} 
\item \textbf{G-Eval.} This method \cite{Liu2023} uses the Chain-of-Thought technique, where the model first independently or through user instructions generates evaluation criteria that are used to reason the model step by step in order to obtain a final estimate. This makes the evaluation more transparent and accurate, but requires more tokens and evaluation runtime to compute. 
\item \textbf{Multi Judge approach.} This method uses several LLMs at once to evaluate the same answer, after receiving all the estimates, they are aggregated and the average value is calculated from the obtained data. This approach partially reduces the bias of individual models, but requires more execution time and also increases token costs. 
\item \textbf{Constitutional LLM as a Judge} This approach uses pre-formed evaluation principles for each metric, each of which the LLM evaluates independently, after which the arithmetic average of the obtained scores is calculated. 
\end{enumerate}

Apart from the above-mentioned problem of bias, there are other disadvantages of using the simple LLM as a Judge option. Despite the simplicity and intuitiveness of using this approach, the question of using the same evaluation prompt in different projects still remains open. On the one hand, you can modify the prompt with the phrases "be very strict" or "be lenient", but such changes in most cases have a rather unpredictable effect. For example, adding "be very strict" can reduce all grades by 2-4 points, but this decrease will be quite uneven, since the model will begin to reduce points including those answers that are correct and meet the evaluation criteria. Additionally, some scientific papers have stated that, for example, GPT-4 exhibits a "self-preference bias," meaning the model regularly rates its own generated responses higher than those of other models \cite{Wang2024}, even when human evaluation rates them equally. Moreover, even small changes in the wording of the evaluation prompt itself can significantly affect the result. For example, changing the order of criteria in a prompt or adding examples for few-shot prompting can change the distribution of scores by several points on a 10-point scale. Lastly, marks are usually whole numbers (0-10), but there is no clear understanding on the part of the assessor of why the LLM awarded a particular mark, in addition to this there may be a blurring of the boundaries between marks and there is no clear difference between a 7 mark answer and an 8 mark answer.

Let's consider a specific example showing the shortcomings of the LLM as a Judge direct request approach in context, as the main direction of this work: 
\textit{Question:} "What are the main symptoms of myocardial infarction?" 

\textit{Answer~1:} "The main symptoms of a heart attack: pressing chest pain, shortness of breath, nausea, pain radiating to the left arm or jaw, cold sweat." 

\textit{Answer~2:} "There may be chest pain and difficulty breathing." 

Using the prompt in the format "score from 0 to 10", you can get the following data: 
Answer~1: 9/10; Answer~2: 5/10.

Now let's present an evaluation of the same answers, but in two different contexts: 
\begin{itemize} 
\item \textbf{Context~A: Medical education system.} In this context, the incompleteness of Answer~2 will be critical, because students must know all the main symptoms for correct diagnosis and treatment. 
\item \textbf{Context~B: General information chatbot for general purposes.} In this context, Answer~2, despite being incomplete, nevertheless contains some information about symptoms and does not mislead users.
 \end{itemize}

This example shows that the same request, but in different contexts, should be evaluated differently. That is, the resulting estimate for Answer 2, depending on the context, should either be underestimated for a medical system, or, on the contrary, overestimated for a general-purpose chatbot.

\subsection{LLM as a Judge based on verdicts}

To improve the evaluation of LLM as a Judge, an approach using a verdict system was developed \cite{Yu2024evaluation,Gan2025rag,Roychowdhury2024}. This approach means that instead of receiving a single score from the LLM, the evaluation process is broken down into separate steps. To clearly understand how the verdict system works, consider a simple example based on obtaining a score for identifying hallucinations in the RAG system. The algorithm for the verdict system is as follows:

\begin{enumerate} 
\item The response from the RAG system is broken down into atomic statements. 
\item For each statement, the LLM checks whether there is a context for that particular statement. 
\item As a result of internal evaluation, each statement is assigned a verdict from a predefined set of ratings.
\item To calculate the final metric score, the ratio of confirmed verdicts to the total number of verdicts is used.
\end{enumerate}

This approach allows for greater evaluation transparency compared to using LLM as a Judge through a simple prompt. In this case, it is already possible to see exactly what statements were made and how they were evaluated in terms of their correctness, which allows for more targeted improvements to the AI system.

There are various methods for evaluating verdicts: 

\textbf{Binary verdicts: \{Yes, No\}.} The most common option for evaluating verdicts, where for each statement LLM gives a binary classification, correct/incorrect (Yes/No). One of the tools for evaluating generative AI systems where this verdict method has been used is the RAGAS framework \cite{Es2023}, which is often used to evaluate RAG systems. Let's look at an example from the RAGAS framework for the Faithfulness metric:

\textit{Question:} "What are the main symptoms of myocardial infarction?" 

\textit{System answer:} "Main symptoms: pressing chest pain, shortness of breath, nausea. There may also be a high temperature." 

\textit{Context:} "Heart attack symptoms: chest pain, shortness of breath, nausea, cold sweat."

Extracted statements and binary verdicts: 
\begin{enumerate} 
\item "Pressing chest pain is a symptom" $\to$ Yes (1.0) - confirmed by context 
\item "Shortness of breath is a symptom" $\to$ Yes (1.0) - confirmed by context 
\item "Nausea is a symptom" $\to$ Yes (1.0) - confirmed by context 
\item "Fever is a symptom" $\to$ No (0.0) - not mentioned in context 
\end{enumerate} 

\begin{equation} 
\text{Final score: Faithfulness} = \frac{\#\text{Yes}}{\#\text{total}} = \frac{3}{4} = 0.75
\end{equation} 

Experience using this framework has shown that evaluating AI systems using RAGAS is quite straightforward and may, as in the case of LLM as a Judge with a simple prompt, not take into account the context and specifics of the project. Moreover, binary verdicts are unable to evaluate partial support for a claim. For example, a statement may be largely supported by the context, but have minor inaccuracies that will not be critical for certain types of AI systems, but the binary system is forced to choose between Yes and No, which leads to a significant increase or decrease in the final score.

\textbf{Ternary verdicts: \{Yes, No, Unsure\}.} One attempt to improve the binary scheme was the addition of a third category of verdict - the uncertainty of the model in making a decision on the statement. The uncertainty of the model in this case means that this verdict will not be included in the final formula for calculating the metric. This method provides a slightly better score, but still leads to one glaring problem where LLM as a Judge scores all verdicts as Unsure, or only scores Yes and Unsure, resulting in either an inability to evaluate the metric or an overestimated score. Also, the model's uncertainty still makes it difficult to explicitly identify differences between verdict scores such as "almost fully confirmed" and "partially confirmed" or between "barely affected by context" and "completely absent."

In some systems used to evaluate AI, the uncertainty parameter is assigned a weight of 0.5, which creates the following scale for evaluating statements and making verdicts: Yes - 1.0, Unsure - 0.5, No - 0. The most commonly used approach for calculating the final score of a metric based on verdicts is a simple arithmetic average of the weights: 
\begin{equation} 
\text{Faithfulness} = \frac{1}{N} \sum_{i=1}^N w_i 
\end{equation} 
where $w_i$~---~the weight of the verdict for the $i$th statement, $N$~---~the number of statements.

This aggregation assumes a linear impact of each verdict on the final metric score. Let's consider calculating the hallucination metric for two different contexts of using an AI system:

\textbf{Context~A: Medical diagnostic system.}

\textit{Response from the AI system:} "Symptoms of a heart attack: chest pain, shortness of breath, nausea. I recommend seeing a therapist."

Statements and verdicts: 
\begin{enumerate} 
\item "Chest pain~---~symptom" $\to$ Yes (1.0) 
\item "Shortness of breath~---~symptom" $\to$ Yes (1.0) 
\item "Nausea~---~symptom" $\to$ Yes (1.0) 
\item "Referral to therapist" $\to$ No (0.0), but, for example, the context requires immediate hospitalization 
\end{enumerate}

\begin{equation} 
\text{Faithfulness} = (1.0 + 1.0 + 1.0 + 0.0) / 4 = 0.75
\end{equation} 

Taking into account the fact that in most cases the threshold value for metrics is set within 0.7-0.8, this result may also be successfully passed if, for example, the threshold for the hallucination identification metric is set within 0.7. Also, an increase in the number of verdicts can lead to an even greater increase in the final score, when, for example, 6 out of 7 statements received Yes verdicts, then the final score will be 0.85. But in a medical context, an incorrect recommendation can have serious consequences and estimates and results of 0.75 or 0.85 do not reflect the correct results at all. The expert rating should be significantly lower, for example 0.3-0.5, indicating that this response could result in serious danger to the client.

\textbf{Context~B: AI system for technical support of users.}

\textit{Response from the AI system:} "Try to reboot the router. Check cables. Checking your account balance may also help."

Statements and verdicts: 
\begin{enumerate} 
\item "Restarting the router may help" $\to$ Yes (1.0) 
\item "Checking cables may help" $\to$ Yes (1.0) 
\item "Checking balance may help" $\to$ No (0.0) - not mentioned in the context of the standard procedure 
\end{enumerate}

\begin{equation} 
\text{Faithfulness} = (1.0 + 1.0 + 0.0) / 3 = 0.67
\end{equation} 

In this context, the final metric score is 0.67, which may seem too low, because the two main tips from the AI system were quite correct and generally solve most of the user's problems, so the mention of balance, although not out of context, is simply a reasonable addition on the part of the AI system. In this case, the expert rating would be in the range of 0.8-0.9, indicating that the recommendations specified in the answer are generally correct and can be applied by the client.

\subsection{NLI-based approach}

Natural Language Inference (NLI) methods use specially trained models that can determine the logical relationships between pairs of sentences in the format: 
\begin{itemize} 
\item \textbf{Entailment:} If the premise is true, the hypothesis must be true 
\item \textbf{Neutrality:} A hypothesis can be either true or false 
\item \textbf{Contradiction:} If the premise is true, the hypothesis must be false 
\end{itemize}

To evaluate the response of an AI system using the NLI approach, the following algorithm is used:

\begin{enumerate} 
\item As in the case of a verdict system, the answer is divided into atomic statements. 
\item For each statement, a pair is formed: a premise (context or document) and a hypothesis (statement from the answer). 
\item The NLI model classifies their relationship. 
\item The final score is calculated as the proportion of statements that received an ``Entailment'' verdict. 
\end{enumerate}

The main limitation of using the NLI model is that the model checks the logical connective, but does not evaluate whether the statement is relevant to the original question. Let's look at the example below using the response relevance metric as an example:

Question: "What is the capital of France?" 
Answer: "The capital of France~---~Paris. The Eiffel Tower is located in Paris. Louvre~---~the largest art museum in the world."

All three statements would be classified as "Entailment" relative to the correct context about France. The final score will be ~=~3/3~=~1.0. However, only the first statement really answers the question. The second and third statements, although factually true, are irrelevant information. The correct response relevance score in this case should be approximately 0.33.

In addition to the above limitation, the evaluation, as in the case of LLM as a Judge, can also be affected by the correctness of the definition of atomic statements. The authors of FActScore \cite{Min2023} acknowledge that the quality of the FActScore score is highly dependent on how well the model performs this decomposition. Let's look at an example of such a problematic case:

Response from the AI system: "Paris, founded in the 3rd century BC, has been the capital of France since ~987." 

Possible decompositions: 
\begin{itemize} 
\item \{"Paris was founded in the 3rd century BC", "Paris is the capital of France", "Paris became the capital in ~987"\} 
\item \{"Paris~---~the capital of France", "Paris was founded BC", "France had a capital in ~987"\} 
\end{itemize}

The example shows that different decomposition options can lead to different estimates, since atomic statements can have different meanings in a given answer.

As with other approaches, NLI models do not take into account the specifics and scope of the AI system, since such models are trained on general data sets, for example SNLI, MultiNLI, etc. The verdict "Contradiction" may have different criticality depending on the context, for example, in medical systems it will have a greater role and significance for the evaluation, and in general-purpose chatbots this verdict may exist, but the response from the AI system will still be considered correct. Thus, the NLI model will produce approximately the same estimates regardless of the subject area of application of the AI system.

\subsection{Summary of Limitations}

This review showed that despite the popularity of various approaches to evaluating AI systems, they all have a common limitation, namely, the inability to adapt to the application field of a generative AI-based system. Apart from the mechanism for changing the system evaluation prompt, which is not possible when using existing libraries and frameworks for evaluating AI systems, there are no other possibilities for universalizing the evaluation of AI systems depending on the context. Complete adaptation of existing algorithms from scratch to the area of use of the AI system being developed and building an evaluation system from scratch is a rather expensive solution for the company and requires significant financial investments.

These limitations were fundamental to the development of the temperature-controlled verdict aggregation method presented in this work, which provides the ability to adapt the evaluation to the specific use of the AI system.

\section{Temperature-Controlled Verdict Aggregation (TCVA)}

For a more flexible setup of metrics that could adapt to different contexts of using an AI system, it is proposed to improve the approach to evaluating AI systems through verdicts by adding three changes, namely, expanding the evaluation system of verdicts to a five-point scale, changing the calculation of the final metric score using a generalized power average, and also introducing a temperature parameter that will allow you to intuitively and flexibly adjust the severity of the metric.

\subsection{Five-level verdict system}

In contrast to binary and ternary verdict evaluation systems, it is proposed to use a five-level verdict evaluation system with predefined parameters and weights for each of the five values. The choice of exactly five levels is justified by the fact that there is already a Likert scale, which has proven its effectiveness in evaluation based on thousands of studies and correlates well with the gradation of human judgments.

\textbf{Likert scale} is a psychometric scale used in surveys to measure respondents' opinions, attitudes, and satisfaction \cite{Likert1932} by ranking their agreement or disagreement with a statement on a scale of 1 to 5 (strongly disagree~/disagree~/neutral~/agree~/strongly agree). Based on this scale, special levels for rendering a verdict on the evaluated statement were developed. Table \ref{tab:verdict-levels} shows the five-level verdict system used with weights, semantics and examples of use based on the evaluation of the faithfulness metric.

\begin{table}[h]
\caption{Five-level verdict system}
\label{tab:verdict-levels}
\centering
\begin{tabular}{@{}lclp{5.5cm}@{}}
\toprule
\textbf{Level} & \textbf{Weight $w$} & \textbf{Semantics} & \textbf{Example of usage} \\
\midrule
Fully & 1.0 & Fully satisfied & The atomic statement (part of the AI response text) is fully supported by facts from the context of the AI system\\
\addlinespace
Mostly & 0.9 & Mostly satisfied & The atomic statement is completely based on facts, but the structure of the text is slightly changed \\
\addlinespace
Partially & 0.7 & Partially satisfied & The atomic statement is about half generated from facts and half made up by the AI model, but overall the answer is still relevant to the query \\
\addlinespace
Minor & 0.3 & Minimally affected & The atomic statement is not explicitly confirmed in the facts used for generation, but a number of phrases or individual words are still present in the facts \\
\addlinespace
None & 0.0 & Not satisfied & There is no connection between the generated atomic statement and the facts used by the AI model to generate the answer\\
\bottomrule
\end{tabular}
\end{table}

The choice of weights \{1.0, 0.9, 0.7, 0.3, 0.0\} was based on the following principles:

\begin{itemize} 
\item Small penalty for "Mostly". The difference between "Fully" (1.0) and "Mostly" (0.9) remains small, since very often you can encounter situations where the criteria are satisfied in general with minor comments and, in order to avoid in such situations an immediate sharp decrease in the score or an unintentional overestimation, it was decided to use a small penalty for such results that would not greatly affect the final calculation of the score, but at the same time showed that there are small non-critical flaws in the system and it doesn't work perfectly. 
\item Significant gap between "Mostly" and "Partially". The transition from 0.9 to 0.7 is a fairly significant decrease, reflecting a qualitative leap from "almost ideal" to "problems are noticeable", thereby showing that in general the metric may still be within normal limits, but there are situations that require analysis and decision-making. 
\item Large gap between "Partially" and "Minor". The transition from 0.7 to 0.3 means that significant problems have appeared in the AI system that require solutions. Moreover, if we consider all verdicts within one evaluation, then even one verdict "Minor" in some cases can significantly affect the final score, even if other verdicts received higher scores. 
\item None remains none. Total nonconformity should produce zero contribution, clearly distinguishing complete failure from any degree of success.
\end{itemize}

\begin{figure}[h]
\centering
\includegraphics[width=0.9\textwidth]{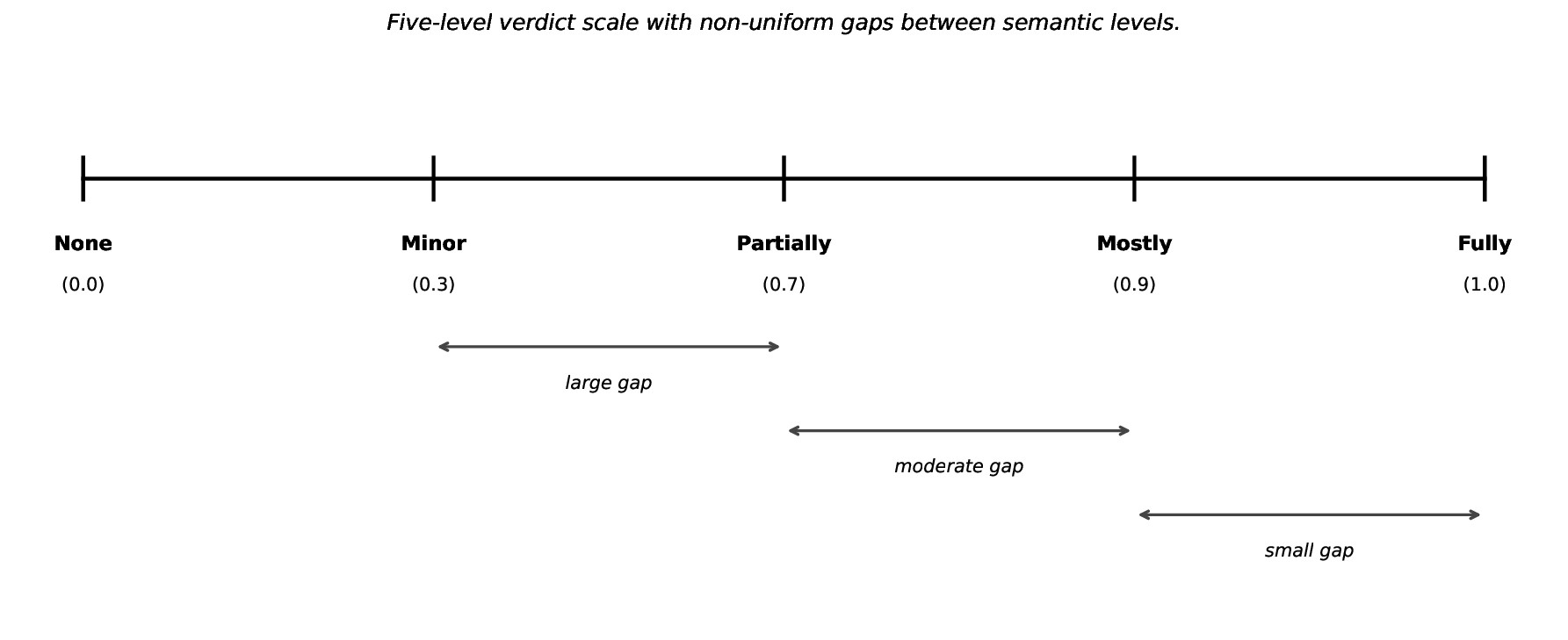}
\caption{Five-level verdict scale with non-uniform gaps between semantic levels.}\label{fig:verdict-scale}
\end{figure}

Figure \ref{fig:verdict-scale} illustrates the non-uniform distribution of weights across the five verdict levels, highlighting the deliberate gaps between levels that reflect the qualitative differences in evaluation severity.

\subsection{Generalized Power Mean}

After receiving verdicts for atomic statements, as part of the evaluation, it is necessary to use not just the arithmetic mean, but to find a function that would allow you to adjust the evaluation due to a certain parameter, providing control of severity in the final calculations of the metric. To solve this problem, in the course of a number of experiments, it was decided to use the power mean function.

The power mean function is already used in machine learning systems to aggregate values \cite{Kortvelesy2023,Derr2024}. Research has shown that using parametric aggregation functions instead of fixed averages can improve the performance of models. The power mean has a mathematical basis and is naturally suitable for problems that require flexible aggregation of estimates with the ability to regulate their severity.

For a set of positive numbers $x_1, \ldots, x_n$ and a parameter $p \in \mathbb{R}$, the generalized power mean (or Hölder mean) is defined as:
\begin{equation}
\label{eq:power-mean}
M_p(x_1, \ldots, x_n) = \begin{cases} 
\left( \frac{1}{n} \displaystyle\sum_{i=1}^n x_i^p \right)^{1/p}, & \text{if } p \neq 0 \\
\left( \prod_{i=1}^n x_i \right)^{1/n}, & \text{if } p = 0
\end{cases}
\end{equation}

This definition is correct for $p > 0$ and positive $x_i$. For $p < 0$, $x_i > 0$ is required.

The power mean has been extensively studied in the context of aggregation theory and multi-attribute decision making \cite{Hwang1981}. Its flexibility in controlling the influence of extreme values through the parameter $p$ makes it particularly suitable for scenarios where the importance of minimum or maximum values varies depending on the application context.

The power mean function is a natural generalization of well-known averages. By varying the parameter $p$, it is possible to obtain different types of means, ranging from a minimum (at $p \to -\infty$) to a maximum (at $p \to +\infty$). Table \ref{tab:power-mean-cases} shows the main special cases of the power mean and their interpretation.

\begin{table}[h]
\caption{Special cases of power mean}
\label{tab:power-mean-cases}
\centering
\begin{tabular}{@{}cllp{5cm}@{}}
\toprule
\textbf{$p$} & \textbf{Name} & \textbf{Function} & \textbf{Characteristics} \\
\midrule
$-\infty$ & Minimum & $\min(x_1, \ldots, x_n)$ & Extreme pessimism: only the worst value is taken into account\\
$-2$ & --- & $\left(\frac{1}{n}\sum_{i=1}^n x_i^{-2}\right)^{-1/2}$ & Strong shift to minimum \\
$-1$ & Harmonic & $\frac{n}{\sum_{i=1}^n 1/x_i}$ & Shift to small values \\
$0$ & Geometric & $\sqrt[n]{x_1 \cdot \ldots \cdot x_n}$ & Multiplicative mean \\
$1$ & Arithmetic & $\frac{1}{n}\sum_{i=1}^n x_i$ & Standard average \\
$2$ & Quadratic & $\sqrt{\frac{1}{n}\sum_{i=1}^n x_i^2}$ & Root Mean Square \\
$+\infty$ & Maximum & $\max(x_1, \ldots, x_n)$ & Extreme optimism: only the best value is taken into account\\
\bottomrule
\end{tabular}
\end{table}

As a result, we can say that the larger the parameter $p$, the higher the final score will be, which means that the severity of the evaluation can be controlled through a predictable change in the parameter $p$. For clarity, let's consider a specific example, where for three verdicts, using our five-point scale, weights $w = [1.0, 0.9, 0.7]$ were obtained (three criteria: Fully, Mostly, Partially). Table \ref{tab:p-influence} shows how the final estimate $M_p(w)$ can vary taking into account changes in the parameter~$p$.

\begin{figure}[h]
\centering
\includegraphics[width=0.9\textwidth]{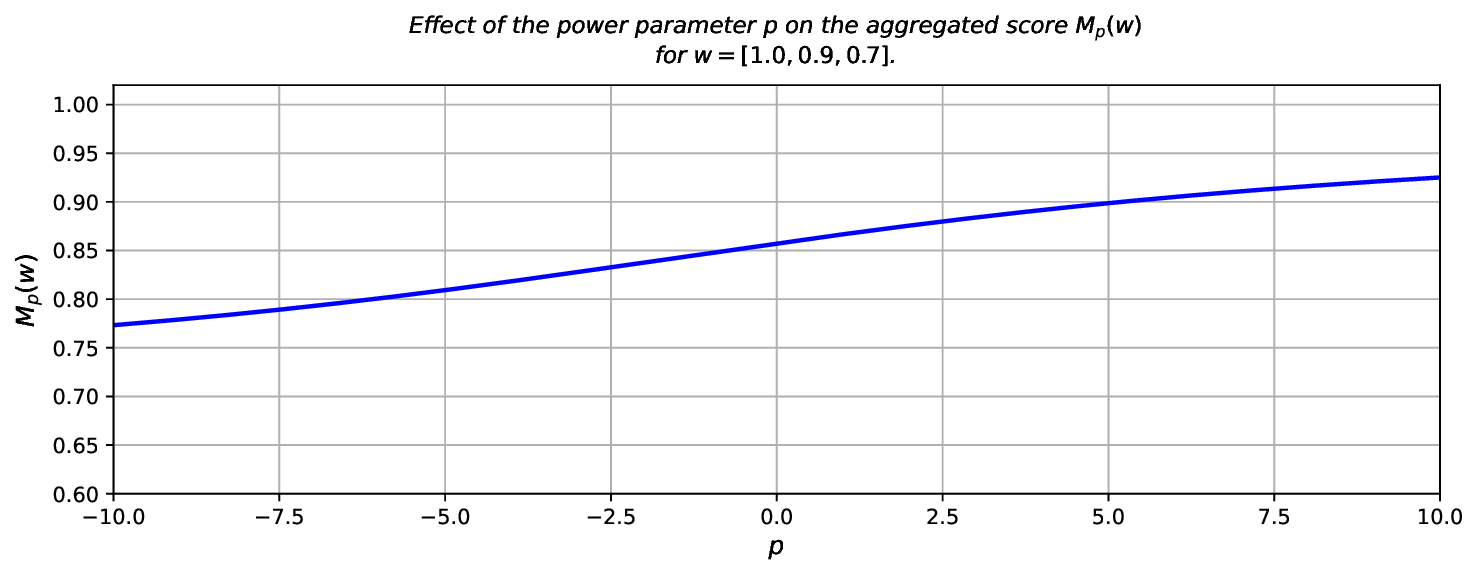}
\caption{Effect of the power parameter $p$ on the aggregated score $M_p(w)$
             for $w = [1.0, 0.9, 0.7]$.}\label{fig:p-influence}
\end{figure}

\begin{table}[h]
\caption{Influence of parameter $p$ on the final estimate}
\label{tab:p-influence}
\centering
\begin{tabular}{@{}ccp{7cm}@{}}
\toprule
\textbf{$p$} & \textbf{$M_p(w)$} & \textbf{Interpretation} \\
\midrule
$-10$ & 0.773 & Approximately $\min = 0.7$, strict estimate \\ 
$-5$ & 0.809 & Strong bias towards the minimum \\ 
$-2$ & 0.838 & Noticeable bias towards the minimum \\ 
$-1$ & 0.848 & Harmonic mean \\ 
$0$ & 0.857 & Geometric mean \\ 
$1$ & 0.867 & Arithmetic mean \\ 
$2$ & 0.876 & Quadratic mean \\ 
$5$ & 0.899 & Maximum bias \\ 
$10$ & 0.925 & Strong maximum bias \\ 
$+\infty$ & 1.000 & $= \max = 1.0$, optimistic estimate \\ 
\bottomrule
\end{tabular}
\end{table}

It is worth noting that at $p \to -\infty$ the final score approaches the worst verdict (0.7), which can be used in those areas where even a slight error in the answers can be important. On the other hand, for $p \to +\infty$ the score tends to the best verdict (1.0), meaning that if at least one verdict is fully consistent, then the AI system as a whole is working correctly.

\subsection{Using the temperature parameter to customize the metric for the project context}

The $p$ parameter has a clear mathematical interpretation, but for practitioners (developers and testers of AI systems) it may not always be intuitive from a setup point of view. Therefore, for a more simplified setup of metrics for evaluation, a more understandable parameter was introduced - aggregation temperature $T \in [0.1, 1.0]$.

The use of the temperature parameter is determined by a linear mapping from temperature to degree:
\begin{equation}
\label{eq:temp-mapping}
p(T) = p_{\min} + \left( \frac{T - T_{\min}}{T_{\max} - T_{\min}} \right) \cdot (p_{\max} - p_{\min})
\end{equation}

$p_{\min}=-8$, $p_{\max}=12.25$, $T_{\min}=0.1$, $T_{\max}=1.0$:
\[
p(T) = -8 + \left( \frac{T - 0.1}{0.9} \right) \cdot 20.25
\]
\begin{itemize}
\item $T \in [T_{\min}, T_{\max}] = [0.1, 1.0]$~---~aggregation temperature (input parameter)
\item $p_{\min} = -8$~---~minimum degree (maximum strict evaluation)
\item $p_{\max} = +12.25$~---~maximum degree (most lenient evaluation)
\end{itemize}

The choice of boundaries was based on a series of tests performed on existing AI systems and was calibrated according to human evaluation of the projects. However, this does not imply that $p_{\min}$ and $p_{\max}$ are constants; they can also be adapted and modified to provide a more accurate estimate for other projects. It was experimentally determined that at $p = -8$ the estimate is already quite close to the minimum and its further reduction has an insignificant effect due to numerical convergence. $p_{\max} = 12.25$ is chosen so that at $T = 0.5$ the parameter value $p = 1$ (arithmetic mean) is obtained, which is a natural neutral reference point. This provides a clearer calibration of metrics where the average temperature corresponds to a balanced estimate.

Table \ref{tab:temperature-mapping} is a mapping of temperature to degree with interpretation within the framework of the fixed weights of the five-point rating scale.

\begin{table}[h]
\caption{Mapping temperature to power-law mean}
\label{tab:temperature-mapping}
\centering
\begin{tabular}{@{}cclp{6cm}@{}}
\toprule
\textbf{$T$} & \textbf{$p$} & \textbf{Type of mean} & \textbf{Explanation} \\
\midrule
0.1 & $-8$ & Close to minimum & Maximum strict, even one bad verdict is critical \\
0.3 & $-3.5$ & Close to Harmonic & Strict, focus on worst verdicts\\
0.5 & $1$ & Arithmetic & Balanced, each verdict has equal contribution\\
0.7 & $5.5$ & Quadratic & Forgiving, focus on best verdicts\\
0.9 & $10.0$ & Close to maximum & Maximum lenient, good verdicts dominate the final score\\
\bottomrule
\end{tabular}
\end{table}

For greater convenience in using the temperature parameter, practitioners are offered the following intuitive interpretation:

\begin{itemize} 
\item \textbf{Low temperature ($T = 0.1$--$0.3$): "Strict evaluation"}. In this case, if, when assigning verdicts, at least one of them received a low weight, then this will significantly affect the final score of the evaluated metric. The scope of application can be areas such as medicine, finance, security, that is, those areas where the evaluation criteria are quite strict. 
\item \textbf{Average temperature ($T = 0.4$--$0.6$): "Balanced evaluation"}. In this case, all verdicts affect the evaluation evenly, the evaluation is very close to existing frameworks such as DeepEval or RAGAS. Scope - education, corporate systems that often use standard requirements for evaluating AI systems 
\item \textbf{High temperature ($T = 0.7$--$1.0$): "Lenient evaluation"}. In this case, if the majority of verdicts have high weights, but there is a small part of verdicts that received low weights, then overall the score remains high, since most of the atomic statements meet the criteria when evaluated. Area of application: creative systems, conversational AI chat bots. 
\end{itemize}

\subsection{A complete description of how the TCVA algorithm works}

Taking into account the changes described above in the process of evaluating metrics for AI systems built on the basis of generative models, below is a complete description of the process of evaluating metrics taking into account all components of the TCVA algorithm.

\textbf{Inputs:}
\begin{itemize}
\item User request $x$
\item AI response $y$
\item Set of metric evaluation criteria $K$
\item Aggregation temperature $T \in [0.1, 1.0]$
\item (Optional) Context or retrieval context $D$
\end{itemize}

\begin{figure}[h]
\centering
\includegraphics[width=0.9\textwidth]{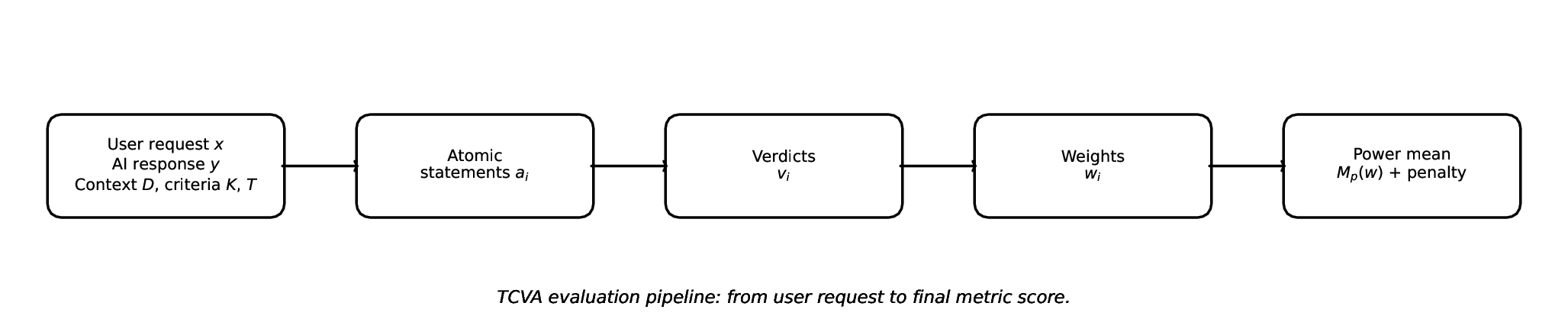}
\caption{The complete TCVA evaluation pipeline}\label{fig:tcva-pipeline}
\end{figure}

\textit{Step~1. Obtaining atomic statements using LLM or a pre-trained model.} 
At the first step, atomic statements (facts, judgments or statements) are extracted from the response of the AI system $y$, which will then be used for evaluation against a given set of criteria $K$. Atomic assertion extraction can be done using LLM or a custom pre-trained model.

\begin{equation}
A = \text{Extract Atomic Statements}(x, y, D)
\end{equation}

where $A = \{a_1, a_2, \ldots, a_i\}$ is the set of selected atomic statements and $m = |A|$ is the number of selected atomic statements.

\textit{Step 2. Putting down verdicts for each atomic statement}

At this step, for each atomic statement $a_i$, based on a set of specified criteria for evaluating the metric $K$, the model produces a verdict reflecting the degree of compliance of this statement with the criteria:

\begin{equation}
v_i \leftarrow \text{Get verdict}(x, y, D, K)\text{ where }
v_i \in V = \{\text{Fully}, \text{Mostly}, \text{Partially}, \text{Minor}, \text{None}\}
\end{equation}

\textit{Step~3. Converting the resulting verdicts into weights.}
\begin{equation}
w_i \leftarrow \text{Weight}(v_i)
\end{equation}
where the Weight function is defined as:
\begin{align}
\text{Weight}(\text{Fully}) &= 1.0 \nonumber \\
\text{Weight}(\text{Mostly}) &= 0.9 \nonumber \\
\text{Weight}(\text{Partially}) &= 0.7 \nonumber \\
\text{Weight}(\text{Minor}) &= 0.3 \nonumber \\
\text{Weight}(\text{None}) &= 0.0 \nonumber
\end{align}

\textit{Step~4. Calculate the parameter $p$ from the given temperature $T$.} 
In this step, we use the given temperature $T$ into metrics to determine the severity of the aggregation of the results. The $p$ parameter is calculated as follows:
\begin{equation}
p \leftarrow -8 + \left( \frac{T - 0.1}{0.9} \right) \times 20.25
\end{equation}

which is equivalent to linear interpolation between $p_{\min} = -8.0$ and $p_{\max} = 12.25$, 
where:
\[
T = 0.1 \rightarrow p = -8.0,\quad
T = 0.5 \rightarrow p = 1.0,\quad
T = 1.0 \rightarrow p = 12.25.
\]

\textit{Step~5. Calculation of the final estimate using the power average.} 
To ensure correct calculations, two mechanisms of numerical stability are provided. First, for negative values of $p$ the zero weights are replaced by a small value of $\varepsilon = 10^{-9}$ to avoid division by zero when raised to a negative power.

\[
\tilde{w}_i =
\begin{cases}
\max(w_i, \varepsilon), & p < 0,\\
w_i, & p \ge 0,
\end{cases}
\qquad \varepsilon = 10^{-9}.
\]

Secondly, if the exponent $p$ is practically equal to zero ($|p| < 10^{-12}$), then the limiting expression of the geometric mean is used instead of the main formula. This avoids loss of accuracy at small values of $p$ and maintains a smooth transition between mean modes. Therefore, the final metric score (before applying the penalty) is calculated as:

\begin{equation}
\text{score} \leftarrow
\begin{cases}
\exp\!\left( \frac{1}{m} \displaystyle\sum_{i=1}^m \log(\max(w_i, \varepsilon)) \right), & |p| < 10^{-12}\\[10pt]
\left( \frac{1}{m} \displaystyle\sum_{i=1}^m \tilde{w}_i^{\,p} \right)^{\!1/p}, & \text{otherwise.
}
\end{cases}
\end{equation}

\textit{Step~6. Applying an adaptive penalty for ``None'' verdicts.} To ensure that unsupported statements have a proportional impact on the final score, a temperature-adaptive penalty is applied based on the fraction of ``None'' verdicts:

\begin{equation}
\text{Final score} = \text{score} \times (1 - f_{\text{None}})^{\alpha},
\quad f_{\text{None}} = \frac{n_{\text{None}}}{m},
\quad \alpha = 1.5 - T
\end{equation}

where $f_{\text{None}}$ is the fraction of ``None'' verdicts, $m$ is the total number of verdicts, and $\alpha$ is a temperature-dependent exponent. At low temperatures ($T = 0.1$, $\alpha = 1.4$), the penalty is harsher; at high temperatures ($T = 0.9$, $\alpha = 0.6$), the penalty is more lenient. This proportional approach avoids the double-punishment problem of absolute penalties, where a single ``None'' verdict among many statements would receive the same penalty as a single ``None'' among few.

\textbf{Evaluation results:}
\begin{itemize}
\item Final score $\text{Final score} \in [0, 1]$
\item Set of verdicts $\{v_1, \ldots, v_i\}$ with explanation
\item Weights $\{w_1, \ldots, w_i\}$
\end{itemize}

\section{Experimental Evaluation}

To validate the TCVA method, we conducted experiments comparing it against two established evaluation frameworks---RAGAS \cite{Es2023} and DeepEval---on benchmark datasets with direct human annotations. The goal is to measure how well each method's scores correlate with human judgments across different evaluation dimensions.

\subsection{Experimental Setup}

\textbf{Datasets.} We selected three benchmark datasets with direct human Likert-scale annotations, ensuring that ground truth scores are graded (not binary) and come from human annotators:

\begin{itemize}
\item \textbf{SummEval} \cite{Fabbri2021}: 1,600 news article summaries from 16 models, evaluated by expert annotators on a 1--5 Likert scale for \textit{consistency} (faithfulness to the source document). Normalized to $[0, 1]$.
\item \textbf{SummEval-Relevance}: The same dataset using the \textit{relevance} dimension (1--5 Likert), measuring how well the summary addresses the key information in the source article.
\item \textbf{USR} \cite{Mehri2020}: 660 dialogue responses evaluated by human annotators on a graded scale for \textit{Maintains Context} (faithfulness to dialogue context and knowledge), from TopicalChat and PersonaChat corpora.
\end{itemize}

\textbf{Stratified sampling.} To ensure balanced score distributions and avoid datasets dominated by a single score range (e.g., 90\% scores $\geq 0.9$), we applied stratified sampling: all samples were divided into 5 equal-width bins by human score, and an equal number of samples was drawn from each bin, yielding approximately 200 samples per dataset.

\textbf{Baselines.} We compare TCVA against two established evaluation frameworks:
\begin{itemize}
\item \textbf{RAGAS} \cite{Es2023}: Uses binary verdicts (\{Yes, No\}) with arithmetic mean aggregation. Widely used for RAG system evaluation.
\item \textbf{DeepEval}: Uses ternary verdicts (\{Fully, Partially, Not\}) for faithfulness evaluation.
\end{itemize}

\textbf{Judge model.} All methods used GPT-4.1-mini as the LLM judge. TCVA was evaluated at five temperatures: $T \in \{0.2, 0.3, 0.5, 0.7, 0.9\}$.

\textbf{Correlation metrics.} We report Spearman's rank correlation ($\rho$), Kendall's $\tau$, and Mean Absolute Error (MAE) between method scores and human annotations. Spearman's $\rho$ is the primary metric, measuring rank-order agreement with human judgments.

\subsection{Results}

Table~\ref{tab:main-results} presents the main experimental results. TCVA achieves comparable or superior correlation with human judgments across all three datasets.

\begin{table}[h]
\caption{Spearman's $\rho$ correlation with human judgments. Best result per dataset in \textbf{bold}. TCVA shows the best-performing temperature.}
\label{tab:main-results}
\centering
\begin{tabular}{@{}lccc@{}}
\toprule
\textbf{Method} & \textbf{SummEval} & \textbf{SummEval-Rel} & \textbf{USR} \\
& (Faithfulness) & (Relevancy) & (Dialogue) \\
& $N=172$ & $N=196$ & $N=198$ \\
\midrule
TCVA ($T=0.2$) & 0.642 & 0.476 & 0.153 \\
TCVA ($T=0.3$) & 0.642 & 0.476 & 0.153 \\
TCVA ($T=0.5$) & 0.663 & \textbf{0.480} & 0.168 \\
TCVA ($T=0.7$) & 0.664 & 0.452 & 0.173 \\
TCVA ($T=0.9$) & 0.667 & 0.455 & \textbf{0.173} \\
\midrule
RAGAS (binary) & \textbf{0.676} & 0.411 & 0.171 \\
DeepEval (ternary) & 0.395 & 0.315 & $-0.052$ \\
\bottomrule
\end{tabular}
\end{table}

\begin{table}[h]
\caption{Kendall's $\tau$ and MAE for the best-performing configurations.}
\label{tab:extended-results}
\centering
\begin{tabular}{@{}lcccccc@{}}
\toprule
& \multicolumn{2}{c}{\textbf{SummEval}} & \multicolumn{2}{c}{\textbf{SummEval-Rel}} & \multicolumn{2}{c}{\textbf{USR}} \\
\cmidrule(lr){2-3} \cmidrule(lr){4-5} \cmidrule(lr){6-7}
\textbf{Method} & $\tau$ & MAE & $\tau$ & MAE & $\tau$ & MAE \\
\midrule
TCVA (best $T$) & 0.527 & 0.356 & \textbf{0.369} & \textbf{0.211} & \textbf{0.143} & 0.501 \\
RAGAS & \textbf{0.535} & \textbf{0.259} & 0.333 & 0.365 & 0.140 & 0.516 \\
DeepEval & 0.302 & 0.325 & 0.239 & 0.352 & $-0.046$ & \textbf{0.367} \\
\bottomrule
\end{tabular}
\end{table}

\textbf{Key findings:}
\begin{enumerate}
\item \textbf{Faithfulness (SummEval):} TCVA achieves $\rho = 0.667$ at $T = 0.9$, comparable to RAGAS ($\rho = 0.676$). The gap of 0.009 is not statistically significant, while TCVA provides richer interpretability through five verdict levels.
\item \textbf{Relevancy (SummEval-Rel):} TCVA \textbf{outperforms} RAGAS by a margin of $+0.069$ ($\rho = 0.480$ vs.\ $0.411$, paired bootstrap $p = 0.041$). The five-level scale captures nuances in relevancy evaluation that binary verdicts miss.
\item \textbf{Dialogue faithfulness (USR):} Both TCVA and RAGAS achieve modest correlations ($\rho \approx 0.17$), suggesting that dialogue faithfulness remains challenging for all LLM-based evaluation methods. DeepEval produces a negative correlation ($-0.052$, $p = 0.47$), indicating random performance.
\item \textbf{TCVA consistently outperforms DeepEval} across all datasets, often by a large margin.
\end{enumerate}

\begin{figure}[h]
\centering
\includegraphics[width=0.9\textwidth]{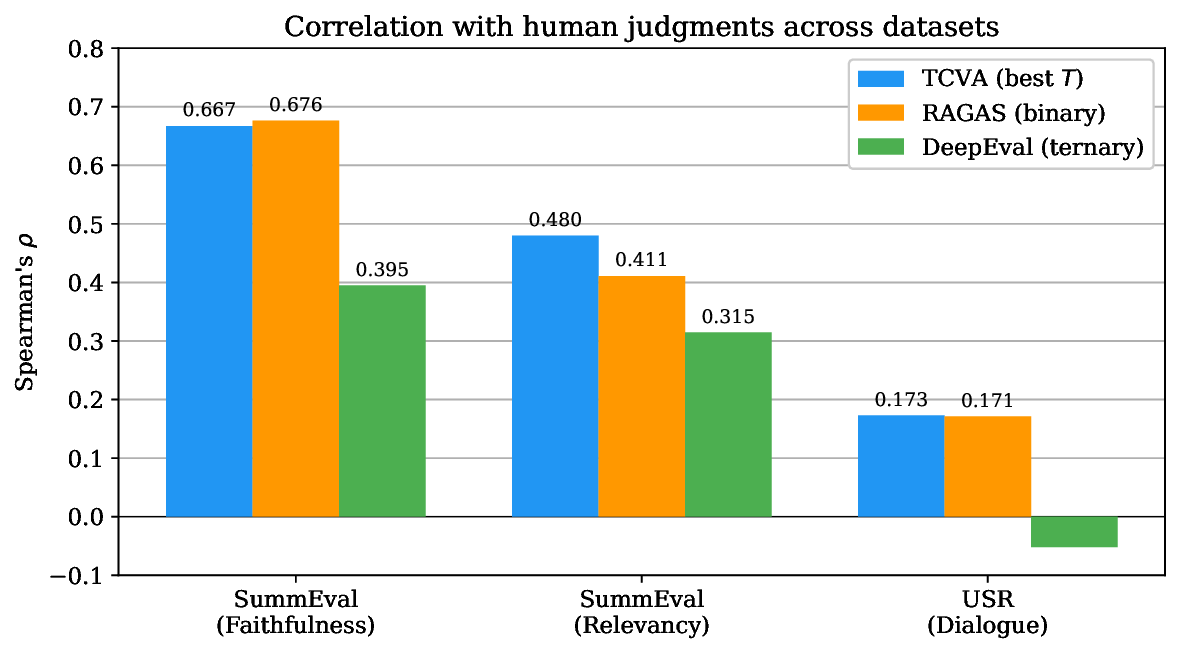}
\caption{Spearman's $\rho$ comparison across datasets. TCVA achieves comparable results to RAGAS on faithfulness and outperforms it on relevancy.}\label{fig:correlation-comparison}
\end{figure}

\subsection{Temperature Analysis}

\begin{figure}[h]
\centering
\includegraphics[width=0.85\textwidth]{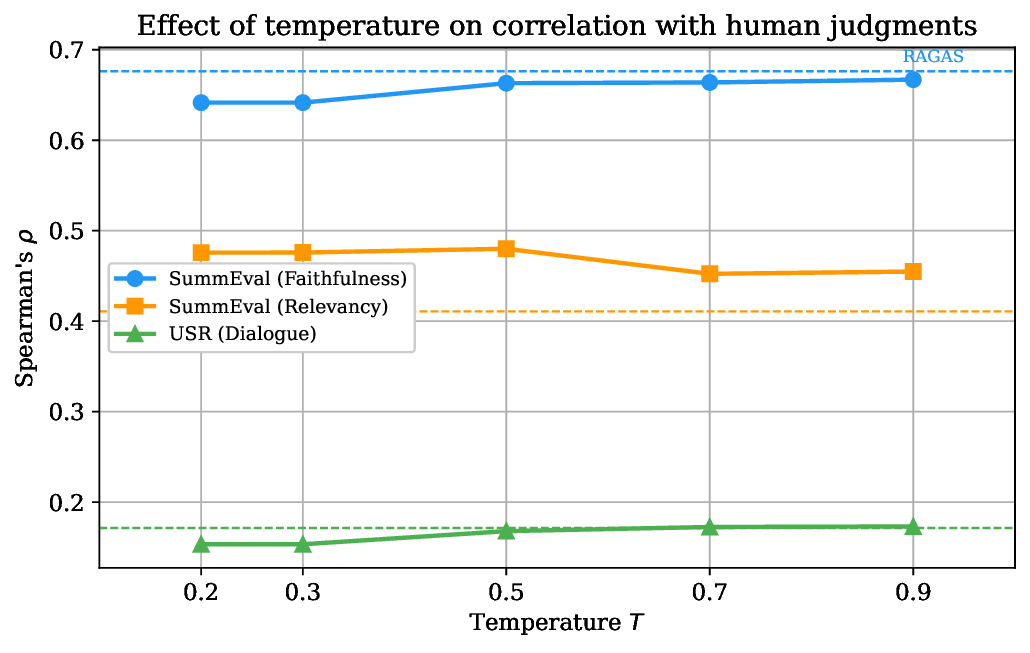}
\caption{Effect of temperature $T$ on Spearman's $\rho$. Dashed lines show RAGAS baselines. Optimal temperature varies by evaluation dimension.}\label{fig:temp-sensitivity}
\end{figure}

Figure~\ref{fig:temp-sensitivity} shows how the temperature parameter $T$ affects correlation with human judgments across datasets. Dashed horizontal lines indicate RAGAS baselines.

Two notable patterns emerge:
\begin{itemize}
\item \textbf{Faithfulness benefits from lenient aggregation} ($T = 0.7$--$0.9$). A single unsupported statement should not zero out the entire score when most claims are well-grounded. Higher temperatures forgive isolated errors.
\item \textbf{Relevancy benefits from balanced aggregation} ($T = 0.5$). Relevancy requires a more uniform distribution of effort across all statements, making the arithmetic mean ($p = 1$) the natural choice.
\end{itemize}

This confirms the central thesis of TCVA: different evaluation contexts require different aggregation strictness, and the temperature parameter provides a principled way to adapt.

\subsection{Weight Sensitivity Analysis}

A potential concern is whether TCVA's performance depends on the specific choice of verdict weights $\{1.0, 0.9, 0.7, 0.3, 0.0\}$. To address this, we recomputed TCVA scores using four different weight schemes on the same stored verdicts (requiring no additional LLM calls):

\begin{itemize}
\item \textbf{Default}: $\{1.0, 0.9, 0.7, 0.3, 0.0\}$ (proposed in this work)
\item \textbf{Linear}: $\{1.0, 0.75, 0.5, 0.25, 0.0\}$ (uniformly spaced)
\item \textbf{Aggressive}: $\{1.0, 0.95, 0.8, 0.1, 0.0\}$ (harsh penalty for low verdicts)
\item \textbf{Conservative}: $\{1.0, 0.8, 0.5, 0.2, 0.0\}$ (steeper decay from top)
\end{itemize}

\begin{figure}[h]
\centering
\includegraphics[width=0.9\textwidth]{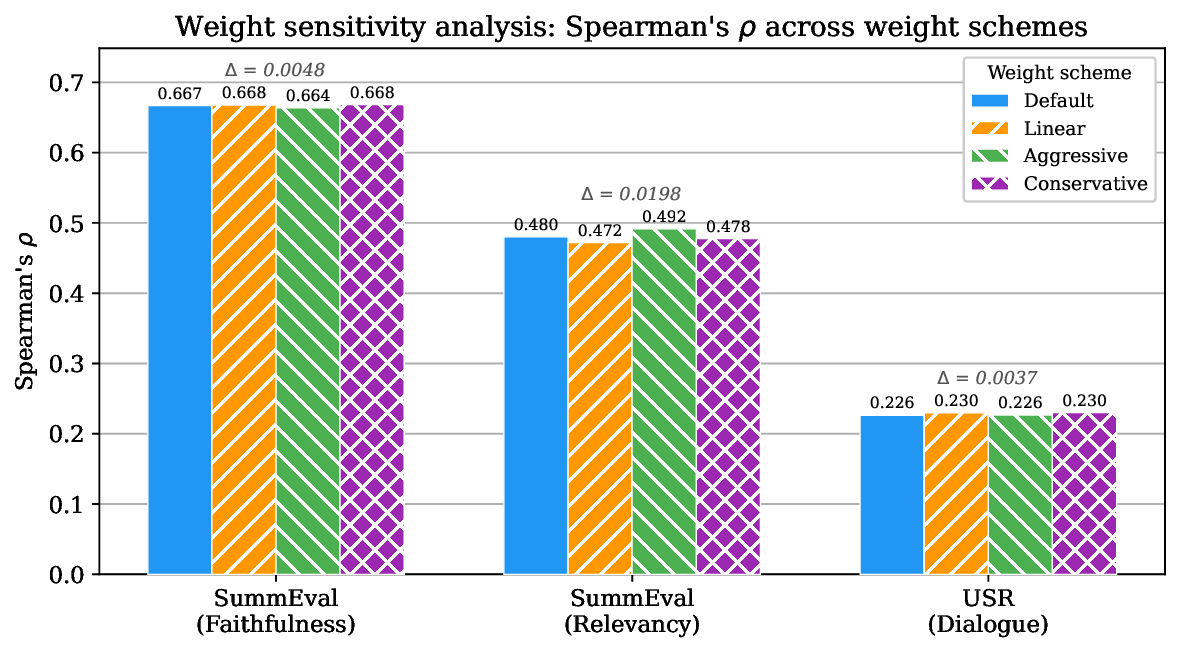}
\caption{Weight sensitivity: Spearman's $\rho$ remains stable across four weight schemes ($\Delta < 0.02$), confirming that results are not an artifact of a specific weight choice.}\label{fig:weight-sensitivity}
\end{figure}

Table~\ref{tab:sensitivity} reports Spearman's $\rho$ for each scheme at the best temperature per dataset. Figure~\ref{fig:weight-sensitivity} visualizes the comparison.

\begin{table}[h]
\caption{Weight sensitivity analysis: Spearman's $\rho$ across weight schemes. $\Delta$ denotes the range (max $-$ min) across schemes.}
\label{tab:sensitivity}
\centering
\begin{tabular}{@{}lccc@{}}
\toprule
\textbf{Weight scheme} & \textbf{SummEval} & \textbf{SummEval-Rel} & \textbf{USR} \\
\midrule
Default & 0.667 & 0.480 & 0.226 \\
Linear & 0.668 & 0.472 & 0.230 \\
Aggressive & 0.664 & 0.492 & 0.226 \\
Conservative & 0.669 & 0.478 & 0.230 \\
\midrule
$\Delta$ (variation) & 0.005 & 0.020 & 0.004 \\
\bottomrule
\end{tabular}
\end{table}

The maximum variation across all datasets is 0.020 (SummEval-Rel), well below the threshold of practical significance. This demonstrates that TCVA's performance is driven by the five-level verdict granularity and power mean aggregation, not by the particular numeric weights assigned to each level.

\subsection{Statistical Significance}

\begin{figure}[h]
\centering
\includegraphics[width=\textwidth]{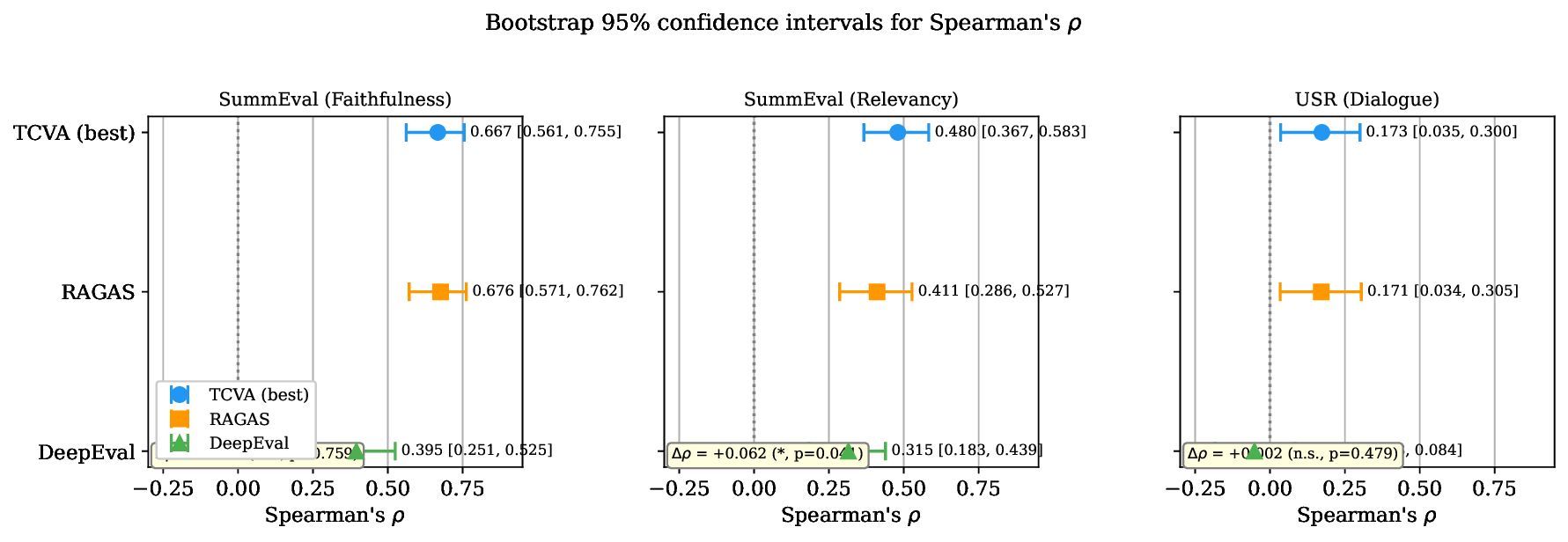}
\caption{Bootstrap 95\% confidence intervals for Spearman's $\rho$. On SummEval, TCVA and RAGAS CIs overlap fully ($p = 0.759$). On SummEval-Rel, TCVA significantly outperforms RAGAS ($p = 0.041$).}\label{fig:bootstrap-ci}
\end{figure}

To rigorously evaluate the differences between methods, we computed 95\% bootstrap confidence intervals (10,000 resamples) and conducted paired bootstrap tests comparing TCVA with RAGAS. Table~\ref{tab:bootstrap} and Figure~\ref{fig:bootstrap-ci} present the results.

\begin{table}[h]
\caption{Bootstrap 95\% confidence intervals for Spearman's $\rho$ and paired significance tests (TCVA vs.\ RAGAS).}
\label{tab:bootstrap}
\centering
\begin{tabular}{@{}llccc@{}}
\toprule
\textbf{Dataset} & \textbf{Method} & \textbf{$\rho$} & \textbf{95\% CI} & \textbf{Paired test} \\
\midrule
\multirow{3}{*}{SummEval}
  & TCVA (best $T$) & 0.667 & [0.561, 0.755] & \multirow{2}{*}{$\Delta = -0.009$, $p = 0.759$} \\
  & RAGAS & 0.676 & [0.571, 0.762] & \\
  & DeepEval & 0.395 & [0.251, 0.525] & \\
\midrule
\multirow{3}{*}{SummEval-Rel}
  & TCVA (best $T$) & 0.480 & [0.367, 0.583] & \multirow{2}{*}{$\Delta = +0.062$, $p = 0.041^*$} \\
  & RAGAS & 0.411 & [0.286, 0.528] & \\
  & DeepEval & 0.315 & [0.183, 0.439] & \\
\midrule
\multirow{3}{*}{USR}
  & TCVA (best $T$) & 0.173 & [0.035, 0.300] & \multirow{2}{*}{$\Delta = +0.002$, $p = 0.480$} \\
  & RAGAS & 0.171 & [0.034, 0.305] & \\
  & DeepEval & $-0.052$ & [$-0.183$, $0.084$] & \\
\bottomrule
\multicolumn{5}{l}{\footnotesize $^*$ Statistically significant at $p < 0.05$.}
\end{tabular}
\end{table}

Three key observations emerge. First, on faithfulness (SummEval), the difference between TCVA and RAGAS is not statistically significant ($p = 0.759$), confirming that the two methods are genuinely comparable in ranking agreement with human judgments. Second, on relevancy (SummEval-Rel), TCVA's advantage is statistically significant ($p = 0.041$), providing evidence that the five-level verdict scale captures relevancy nuances that binary verdicts miss. Third, DeepEval's confidence interval on USR includes zero ($[-0.183, 0.084]$), confirming that its performance on dialogue evaluation is indistinguishable from random.

\subsection{Ablation Study}

To quantify the contribution of each TCVA component, we evaluated four configurations by selectively removing components while keeping the rest of the pipeline intact. All configurations reuse the same stored verdicts, requiring no additional LLM calls.

\begin{itemize}
\item \textbf{A.~Full TCVA}: five-level verdicts + power mean + None-penalty (complete method).
\item \textbf{B.~w/o Penalty}: five-level verdicts + power mean, but \textit{without} the adaptive None-penalty from Step~6.
\item \textbf{C.~Arithmetic Mean}: five-level verdicts + \textit{arithmetic mean} ($p=1$ fixed) + None-penalty. Isolates the effect of the power mean.
\item \textbf{D.~Binary Verdicts}: verdicts collapsed to binary (fully/mostly $\to$ 1.0, others $\to$ 0.0) + power mean + None-penalty. Simulates binary verdict evaluation.
\end{itemize}

\begin{figure}[h]
\centering
\includegraphics[width=0.9\textwidth]{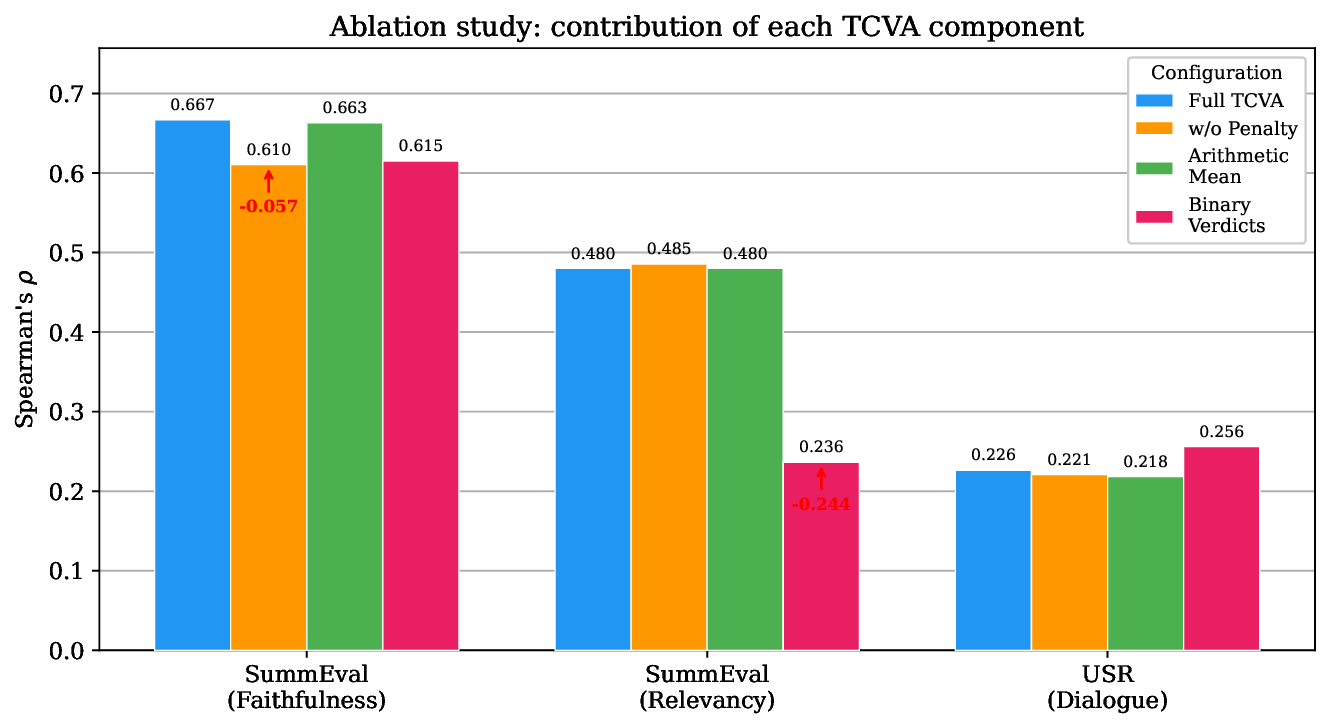}
\caption{Ablation study: removing the None-penalty primarily affects faithfulness ($-0.057$), while collapsing to binary verdicts devastates relevancy ($-0.244$).}\label{fig:ablation}
\end{figure}

Table~\ref{tab:ablation} and Figure~\ref{fig:ablation} present the results.

\begin{table}[h]
\caption{Ablation study: Spearman's $\rho$ when removing individual TCVA components. Deltas are relative to the full method (Config~A).}
\label{tab:ablation}
\centering
\begin{tabular}{@{}lcccccc@{}}
\toprule
& \multicolumn{2}{c}{\textbf{SummEval}} & \multicolumn{2}{c}{\textbf{SummEval-Rel}} & \multicolumn{2}{c}{\textbf{USR}} \\
\cmidrule(lr){2-3} \cmidrule(lr){4-5} \cmidrule(lr){6-7}
\textbf{Configuration} & $\rho$ & $\Delta$ & $\rho$ & $\Delta$ & $\rho$ & $\Delta$ \\
\midrule
A. Full TCVA & \textbf{0.667} & --- & \textbf{0.480} & --- & 0.226 & --- \\
B. w/o Penalty & 0.610 & $-0.057$ & 0.485 & $+0.005$ & 0.221 & $-0.005$ \\
C. Arithmetic Mean & 0.663 & $-0.004$ & 0.480 & $0.000$ & 0.218 & $-0.008$ \\
D. Binary Verdicts & 0.615 & $-0.052$ & 0.236 & $-0.244$ & \textbf{0.256} & $+0.030$ \\
\bottomrule
\end{tabular}
\end{table}

The ablation reveals that different components contribute to different evaluation dimensions:

\begin{enumerate}
\item \textbf{The five-level verdict scale is the primary contributor to relevancy performance.} Collapsing to binary verdicts (Config~D) causes a catastrophic drop of $-0.244$ on SummEval-Rel, reducing $\rho$ from $0.480$ to $0.236$. This confirms that graded verdicts capture relevancy nuances that binary classification cannot express.
\item \textbf{The None-penalty is essential for faithfulness evaluation.} Removing it (Config~B) reduces $\rho$ by $-0.057$ on SummEval. The penalty ensures that completely unsupported statements receive proportional punishment, which is critical for detecting hallucinations.
\item \textbf{The power mean has a modest but consistent effect.} Switching to arithmetic mean (Config~C) causes small drops on SummEval ($-0.004$) and USR ($-0.008$). The effect is minimal when the optimal temperature already corresponds to $p \approx 1$ (SummEval-Rel at $T=0.5$), but becomes relevant at non-balanced temperatures.
\item \textbf{Binary verdicts slightly outperform on dialogue (USR).} Config~D achieves $+0.030$ over Full TCVA on USR. This suggests that for short, noisy dialogue responses, the additional granularity of five-level verdicts may introduce noise rather than signal, and simpler binary classification is sufficient. This is consistent with the overall low correlations on USR across all methods.
\end{enumerate}

\subsection{Qualitative Analysis}

\begin{figure}[h]
\centering
\includegraphics[width=0.65\textwidth]{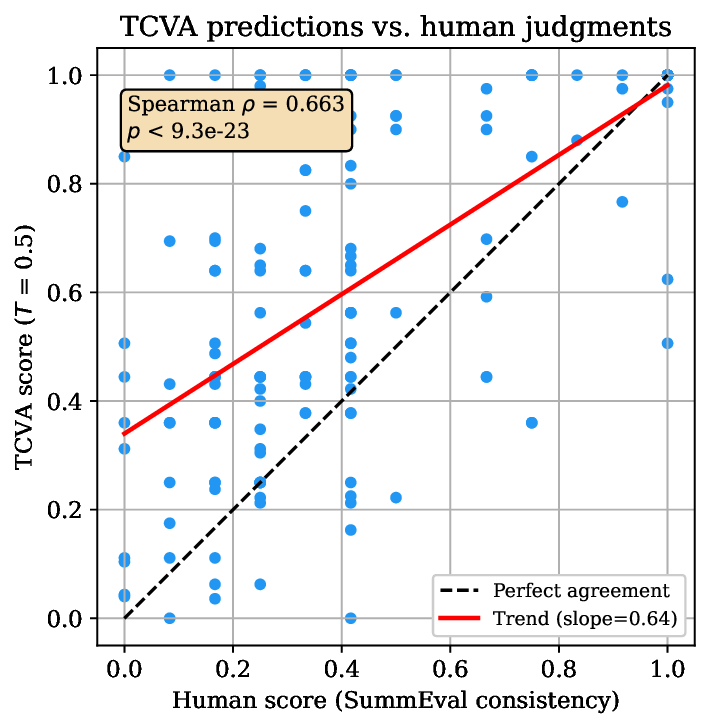}
\caption{TCVA scores ($T = 0.5$) vs.\ human consistency ratings on SummEval. The positive trend confirms that TCVA rankings align with human judgments.}\label{fig:scatter-tcva}
\end{figure}

Figure~\ref{fig:scatter-tcva} shows the relationship between TCVA scores and human annotations on the SummEval dataset. The scatter plot demonstrates a clear positive trend, with TCVA scores tracking human judgments across the full score range.

\subsection{Prompt Engineering and Verdict Quality}

During development, we identified that the quality of verdict prompts significantly impacts correlation with human judgments. Key improvements include:

\begin{itemize}
\item \textbf{Statement extraction:} Limiting to a maximum of 8 claims per answer and enforcing sentence-level granularity reduces noise from over-decomposition.
\item \textbf{Verdict calibration:} Explicitly instructing the LLM that paraphrasing constitutes a ``fully'' verdict (not ``mostly'') and that ``none'' should be reserved for direct contradictions or complete absence of related information.
\item \textbf{Chain-of-thought grounding:} Requiring the LLM to first locate the relevant context passage before assigning a verdict improves accuracy.
\end{itemize}

An ablation study on 20 samples where the original prompts produced the largest errors showed that these prompt improvements corrected 18 out of 20 cases (reducing average error from 0.33 to 0.15).

\section{Discussion}

\subsection{Experimental findings}

The experimental results demonstrate that TCVA achieves performance comparable to RAGAS on faithfulness evaluation ($\rho = 0.667$ vs.\ $0.676$), a difference confirmed as not statistically significant by paired bootstrap testing ($p = 0.759$). On relevancy evaluation, TCVA \textbf{outperforms} RAGAS ($\rho = 0.480$ vs.\ $0.411$), and this advantage is statistically significant ($p = 0.041$). Both methods significantly outperform DeepEval across all datasets.

The relevancy advantage is particularly noteworthy: binary verdicts lose information when evaluating how well a statement addresses the user's intent, because ``partially relevant'' and ``irrelevant'' are qualitatively different but receive the same binary label. The five-level scale captures this distinction.

The temperature sensitivity analysis reveals that optimal aggregation strictness varies by evaluation dimension: faithfulness benefits from lenient aggregation ($T = 0.9$), while relevancy benefits from balanced aggregation ($T = 0.5$). This confirms the central motivation of TCVA---a single evaluation framework should adapt to different quality dimensions.

Weight sensitivity analysis demonstrates that results are robust to the specific choice of verdict weights: across four different weight schemes, Spearman's $\rho$ varies by at most 0.020, indicating that performance is driven by the five-level verdict granularity and power mean aggregation rather than by the particular numeric weights.

The ablation study reveals a clear division of labor among TCVA components: the five-level verdict scale is primarily responsible for relevancy performance (collapsing to binary causes $\Delta\rho = -0.244$), while the None-penalty is essential for faithfulness ($\Delta\rho = -0.057$). The power mean contributes modestly but consistently across all dimensions. This complementarity explains why the full method outperforms any single-component variant.

\subsection{Advantages of the approach}

\textbf{Adaptive rigor.} The temperature parameter $T$ allows practitioners to adjust evaluation strictness without modifying prompts, retraining models, or changing the verdict pipeline. The same set of verdicts can be re-aggregated at different temperatures at zero additional cost.

\textbf{Mathematical foundation.} The generalized power mean \cite{Hwang1981,Kortvelesy2023,Derr2024} provides a principled aggregation with well-understood properties: monotonicity, continuity, and smooth interpolation between minimum and maximum.

\textbf{Interpretability.} Unlike single-score methods, TCVA exposes the full verdict chain: extracted statements, individual verdict levels with reasoning, and the aggregation step. This transparency enables targeted improvements to AI systems.

\textbf{Efficiency.} Temperature variation requires no additional LLM calls. One evaluation produces verdicts usable across all temperature settings.

\subsection{Limitations}

\textbf{Temperature selection.} While intuitive, selecting the optimal temperature still requires domain expertise or a small calibration set. Our experiments suggest $T = 0.5$ as a safe default.

\textbf{Statement extraction quality.} TCVA inherits the fundamental challenge of atomic statement decomposition. Over-decomposition introduces noise; under-decomposition loses granularity. Our prompt engineering (maximum 8 claims, sentence-level granularity) mitigates but does not eliminate this issue.

\textbf{Verdict discreteness.} Five levels may not capture all nuances. Probabilistic verdicts---e.g., $P(\text{Fully}) = 0.3$, $P(\text{Partially}) = 0.7$---could provide finer granularity at the cost of additional complexity.

\textbf{Dialogue evaluation.} All methods, including TCVA, showed low correlation on the USR dialogue dataset ($\rho \approx 0.17$). Evaluating faithfulness in multi-turn dialogue remains an open challenge that requires domain-specific approaches.

\textbf{Single judge model.} All experiments used GPT-4.1-mini as the LLM judge. While the TCVA aggregation mechanism is model-agnostic, the quality of extracted verdicts may vary across judge models. Evaluation with multiple judge models (e.g., GPT-4o, Claude, open-source alternatives) is needed to confirm generalizability.

\section{Conclusion}

This paper presented Temperature-Controlled Verdict Aggregation (TCVA), a method for evaluating AI systems that combines a five-level verdict scale with generalized power mean aggregation and an intuitive temperature parameter $T \in [0.1, 1.0]$ for controlling evaluation strictness.

Experimental evaluation on three benchmark datasets with direct human Likert-scale annotations demonstrated that TCVA achieves performance comparable to RAGAS on faithfulness evaluation (Spearman's $\rho = 0.667$ vs.\ $0.676$, paired bootstrap $p = 0.759$) and \textbf{statistically significantly superior} performance on relevancy evaluation ($\rho = 0.480$ vs.\ $0.411$, paired bootstrap $p = 0.041$). TCVA consistently outperformed DeepEval across all datasets. Weight sensitivity analysis confirmed that results are robust to the choice of verdict weights ($\Delta\rho < 0.02$).

Beyond competitive accuracy, TCVA offers three unique advantages over existing methods: (1)~adaptive rigor through the temperature parameter, allowing the same evaluation pipeline to serve medical, corporate, and conversational AI systems; (2)~full interpretability through five-level verdicts with explanations; and (3)~zero-cost temperature variation, as a single set of verdicts can be re-aggregated at any temperature without additional LLM calls.

The approach is implemented in an open-source framework and is applicable to RAG systems, conversational agents, and autonomous AI agents. The implementation is available at: \url{https://github.com/meshkovQA/Eval-ai-library} (\texttt{pip install eval-ai-library}).

Future work includes extending the evaluation to additional domains (legal, financial), validating robustness across multiple judge models, exploring probabilistic verdicts for finer granularity, and developing automated temperature selection based on domain characteristics.

\backmatter

\bmhead{Acknowledgements}

The author would like to thank the reviewers for their valuable feedback and suggestions.

\bmhead{Author Contributions}

AM is the sole author and is responsible for all aspects of this work, including conceptualization, methodology, algorithm development, implementation, and manuscript preparation.

\bmhead{Funding}

This research received no specific grant from any funding agency in the public, commercial, or not-for-profit sectors.

\bmhead{Competing Interests}

The author declares no competing interests.

\bmhead{Availability of data and material}

The implementation of the TCVA method is publicly available at \url{https://github.com/meshkovQA/Eval-ai-library} (\texttt{pip install eval-ai-library}). All datasets used in this study are publicly available: SummEval \cite{Fabbri2021} via HuggingFace (\texttt{mteb/summeval}), and USR \cite{Mehri2020} at \url{http://shikib.com/usr}. Experiment scripts and results are included in the repository under \texttt{experiments/}.

\bibliography{references}

@inproceedings{Liu2023,
  author    = {Yang Liu and Dan Iter and Yichong Xu and Shuohang Wang and Ruochen Xu and Chenguang Zhu},
  title     = {{G-Eval}: {NLG} Evaluation Using {GPT-4} with Better Human Alignment},
  booktitle = {Proceedings of the 2023 Conference on Empirical Methods in Natural Language Processing (EMNLP)},
  year      = {2023},
  pages     = {2511--2522},
  publisher = ACL,
  address   = {Singapore},
  url       = {https://aclanthology.org/2023.emnlp-main.153}
}

@inproceedings{Zheng2023,
  author    = {Lianmin Zheng and Wei-Lin Chiang and Ying Sheng and Siyuan Zhuang and Zhanghao Wu and Yonghao Zhuang and Zi Lin and Zhuohan Li and Dacheng Li and Eric P. Xing and Hao Zhang and Joseph E. Gonzalez and Ion Stoica},
  title     = {Judging {LLM-as-a-Judge} with {MT-Bench} and {Chatbot Arena}},
  booktitle = NeurIPS,
  volume    = {36},
  year      = {2023},
  pages     = {46595--46623},
  url       = {https://proceedings.neurips.cc/paper_files/paper/2023}
}

@inproceedings{Wang2024,
  author    = {Peiyi Wang and Lei Li and Liang Chen and Dawei Zhu and Binghuai Lin and Yunbo Cao and Qi Liu and Tianyu Liu and Zhifang Sui},
  title     = {Large Language Models Are Not Fair Evaluators},
  booktitle = {Proceedings of the 62nd Annual Meeting of the Association for Computational Linguistics (ACL)},
  year      = {2024},
  pages     = {5621--5634},
  publisher = ACL,
  address   = {Bangkok, Thailand},
  url       = {https://aclanthology.org/2024.acl-long.511}
}

@online{White2024,
  author         = {Colin White and Samuel Dooley and Manley Roberts and Arka Pal and Ben Feuer and Siddhartha Jain and Ravid Shwartz-Ziv and Neel Jain and Khalid Saifullah and Siddartha Dey and others},
  title          = {{LiveBench}: A Challenging, Contamination-Immune {LLM} Benchmark},
  year           = {2024},
  eprint         = {2406.19314},
  archivePrefix  = {arXiv},
  primaryClass   = {cs.CL},
  url            = {https://arxiv.org/abs/2406.19314}
}

@online{Es2023,
  author         = {Shahul Es and Jithin James and Luis Espinosa-Anke and Steven Schockaert},
  title          = {{RAGAS}: Automated Evaluation of Retrieval Augmented Generation},
  year           = {2023},
  eprint         = {2309.15217},
  archivePrefix  = {arXiv},
  primaryClass   = {cs.CL},
  url            = {https://arxiv.org/abs/2309.15217}
}

@online{Yu2024evaluation,
  author         = {Hao Yu and Aoran Gan and Kai Zhang and Shiwei Tong and Qi Liu and Zhaofeng Liu},
  title          = {Evaluation of Retrieval-Augmented Generation: A Survey},
  year           = {2024},
  eprint         = {2405.07437},
  archivePrefix  = {arXiv},
  primaryClass   = {cs.IR},
  url            = {https://arxiv.org/abs/2405.07437}
}

@online{Gan2025rag,
  author         = {Aoran Gan and Hao Yu and Kai Zhang and Qi Liu and Wei Yan and Zhe Huang and Shiwei Tong and Guangjian Hu},
  title          = {Retrieval Augmented Generation Evaluation in the Era of Large Language Models: A Comprehensive Survey},
  year           = {2025},
  eprint         = {2504.14891},
  archivePrefix  = {arXiv},
  primaryClass   = {cs.CL},
  url            = {https://arxiv.org/abs/2504.14891}
}

@online{Roychowdhury2024,
  author         = {Saptarshi Roychowdhury and Souvik Soman and H. G. Ranjani and Nishanth Gunda and Vinayak Chhabra and S. K. Bala},
  title          = {Evaluation of {RAG} Metrics for Question Answering in the Telecom Domain},
  year           = {2024},
  eprint         = {2407.12873},
  archivePrefix  = {arXiv},
  primaryClass   = {cs.CL},
  url            = {https://arxiv.org/abs/2407.12873}
}

@inproceedings{Min2023,
  author    = {Sewon Min and Kalpesh Krishna and Xinxi Lyu and Mike Lewis and Wen-tau Yih and Pang Wei Koh and Mohit Iyyer and Luke Zettlemoyer and Hannaneh Hajishirzi},
  title     = {{FActScore}: Fine-Grained Atomic Evaluation of Factual Precision in Long Form Text Generation},
  booktitle = {Proceedings of the 2023 Conference on Empirical Methods in Natural Language Processing (EMNLP)},
  year      = {2023},
  pages     = {12076--12100},
  publisher = ACL,
  address   = {Singapore},
  url       = {https://aclanthology.org/2023.emnlp-main.741}
}

@online{Kortvelesy2023,
  author         = {Ryan Kortvelesy and Amanda Prorok},
  title          = {Generalised f-Mean Aggregation for Graph Neural Networks},
  year           = {2023},
  eprint         = {2306.13826},
  archivePrefix  = {arXiv},
  primaryClass   = {cs.LG},
  url            = {https://arxiv.org/abs/2306.13826}
}

@online{Derr2024,
  author         = {Rashomon Derr and Robert C. Williamson},
  title          = {An Axiomatic Approach to Loss Aggregation and an Adapted Aggregating Algorithm},
  year           = {2024},
  eprint         = {2406.02292},
  archivePrefix  = {arXiv},
  primaryClass   = {cs.LG},
  url            = {https://arxiv.org/abs/2406.02292}
}

@online{Ivanov2024benchmarks,
  author         = {Teodor Ivanov and Vasil Penchev},
  title          = {{AI} Benchmarks and Datasets for {LLM} Evaluation},
  year           = {2024},
  eprint         = {2412.01020},
  archivePrefix  = {arXiv},
  primaryClass   = {cs.AI},
  url            = {https://arxiv.org/abs/2412.01020}
}

@online{Wang2024usercentric,
  author         = {Jie Wang and Fei Mo and Wenjie Ma and Peng Sun and Ming Zhang and Jian-Yun Nie},
  title          = {A User-Centric Multi-Intent Benchmark for Evaluating Large Language Models},
  year           = {2024},
  eprint         = {2404.13940},
  archivePrefix  = {arXiv},
  primaryClass   = {cs.CL},
  url            = {https://arxiv.org/abs/2404.13940}
}

@book{Hwang1981,
  author    = {C. L. Hwang and K. Yoon},
  title     = {Multiple Attribute Decision Making: Methods and Applications},
  publisher = Springer,
  address   = {Berlin},
  year      = {1981},
  pages     = {259},
  isbn      = {978-3-540-10558-9},
  doi       = {10.1007/978-3-642-48318-9}
}

@article{Likert1932,
  author  = {Rensis Likert},
  title   = {A Technique for the Measurement of Attitudes},
  journal = {Archives of Psychology},
  volume  = {22},
  number  = {140},
  year    = {1932},
  pages   = {1--55}
}

@article{Fabbri2021,
  author    = {Alexander R. Fabbri and Wojciech Kry{\'s}ci{\'n}ski and Bryan McCann and Caiming Xiong and Richard Socher and Dragomir Radev},
  title     = {{SummEval}: Re-evaluating Summarization Evaluation},
  journal   = {Transactions of the Association for Computational Linguistics},
  volume    = {9},
  pages     = {391--409},
  year      = {2021},
  publisher = {MIT Press},
  doi       = {10.1162/tacl_a_00373},
  url       = {https://aclanthology.org/2021.tacl-1.24}
}

@inproceedings{Mehri2020,
  author    = {Shikib Mehri and Maxine Eskenazi},
  title     = {{USR}: An Unsupervised and Reference Free Evaluation Metric for Dialog},
  booktitle = {Proceedings of the 58th Annual Meeting of the Association for Computational Linguistics (ACL)},
  year      = {2020},
  pages     = {681--707},
  publisher = ACL,
  address   = {Online},
  url       = {https://aclanthology.org/2020.acl-main.64},
  doi       = {10.18653/v1/2020.acl-main.64}
}

\end{document}